\documentclass[conference]{IEEEtran}
\IEEEoverridecommandlockouts
\usepackage{url}
\usepackage{dblfloatfix}
\usepackage{cite}
\usepackage{amsmath,amssymb,amsfonts}
\usepackage{algorithmic}
\usepackage{graphicx}
\usepackage{array}
\usepackage{hyperref}

\usepackage{textcomp}
\usepackage{xcolor}
\usepackage{xcolor}
\usepackage{pifont}
\usepackage{makecell}
\usepackage{dblfloatfix}
\usepackage{multirow}
\usepackage[caption=false,font=footnotesize]{subfig}

\newcommand{\cmark}{\textcolor{green!50!black}{\ding{51}}}
\newcommand{\xmark}{\textcolor{red!75!black}{\ding{55}}}

\def\BibTeX{{\rm B\kern-.05em{\sc i\kern-.025em b}\kern-.08em
    T\kern-.1667em\lower.7ex\hbox{E}\kern-.125emX}}
\begin{document}

\title{ChildGaze: A Benchmark Dataset for Collaborative Behavior Understanding in Children\\
\thanks{* These authors contributed equally to this work}
}

\author{
\IEEEauthorblockN{
Sindhuja Penchala\textsuperscript{1,*},
Saketh Reddy Kontham\textsuperscript{1,*},
Prachi Bhattacharjee\textsuperscript{1},
S. Nima Mahmoodi\textsuperscript{1},
Daniel Fonseca\textsuperscript{1},\\
Sareh Karami\textsuperscript{2},
Mehdi Garemani\textsuperscript{2},
Sudip Mittal\textsuperscript{1},
Shahram Rahimi\textsuperscript{1},
Noorbakhsh Amiri Golilarz\textsuperscript{1}
}

\IEEEauthorblockA{
\textsuperscript{1}The University of Alabama, Tuscaloosa, AL, USA, \\
\textsuperscript{2}Mississippi State University, Mississippi State, MS, USA \\
\{spenchala, skontham, pbhattacharjee1\}@crimson.ua.edu,
\{NMahmoodi, Dfonseca\}@eng.ua.edu \\
 skarami@colled.msstate.edu,
 m.ghahremani@msstate.edu, \{sudip.mittal, srahimi1, namirigolilarz\}@ua.edu
}

}

 \maketitle

\begin{abstract}
Understanding collaborative behavior in children is important for analyzing social participation, peer interaction, shared attention, and engagement during play and learning activities. Reliable recognition of these cues can support research in child development, educational analysis, and human-centered computer vision. However, estimating where a child is looking does not necessarily reveal whether the child is actively participating in a shared activity. To support this higher-level analysis, we introduce \href{https://www.kaggle.com/datasets/penchalasindhuja/childgaze-a-benchmark-dataset}{ChildGaze}, a child-centered behavioral annotation dataset built on the ChildPlay video collection\cite{tafasca2023childplay}. ChildGaze introduces two behavioral labels, \emph{collaborative} and \emph{non-collaborative}, assigned independently to each child within a frame. The dataset provides face, left-hand, and right-hand bounding boxes for children and adults and organizes the annotations at the row, person, and frame levels. The current release contains 27 annotated video files, 10,641 frames, 73,268 body-part annotation rows. Annotation reliability was evaluated on 1,187 frames using independent annotations from two annotators. The collaboration labels achieved 93.16\% raw agreement and a Cohen's $\kappa$ of 0.8631, while bounding-box annotations achieved an overall mean IoU of 0.808. Baseline experiments with pretrained ViT and Swin Transformer models achieved up to 97.44\% child-person-level accuracy and 96.80\% frame-level accuracy, respectively. These results show that ChildGaze provides a reliable benchmark for studying collaborative behavior in naturalistic child-adult and peer interactions.

\end{abstract}


\section{Introduction}

\begin{figure*}[!t]
\centering
\subfloat[\textbf{Child~\#1 Collaboration: Yes} (child and adult looking at the
same object); \textbf{Frame Collaboration: Yes} (majority Yes).]{%
    \includegraphics[width=0.32\textwidth]{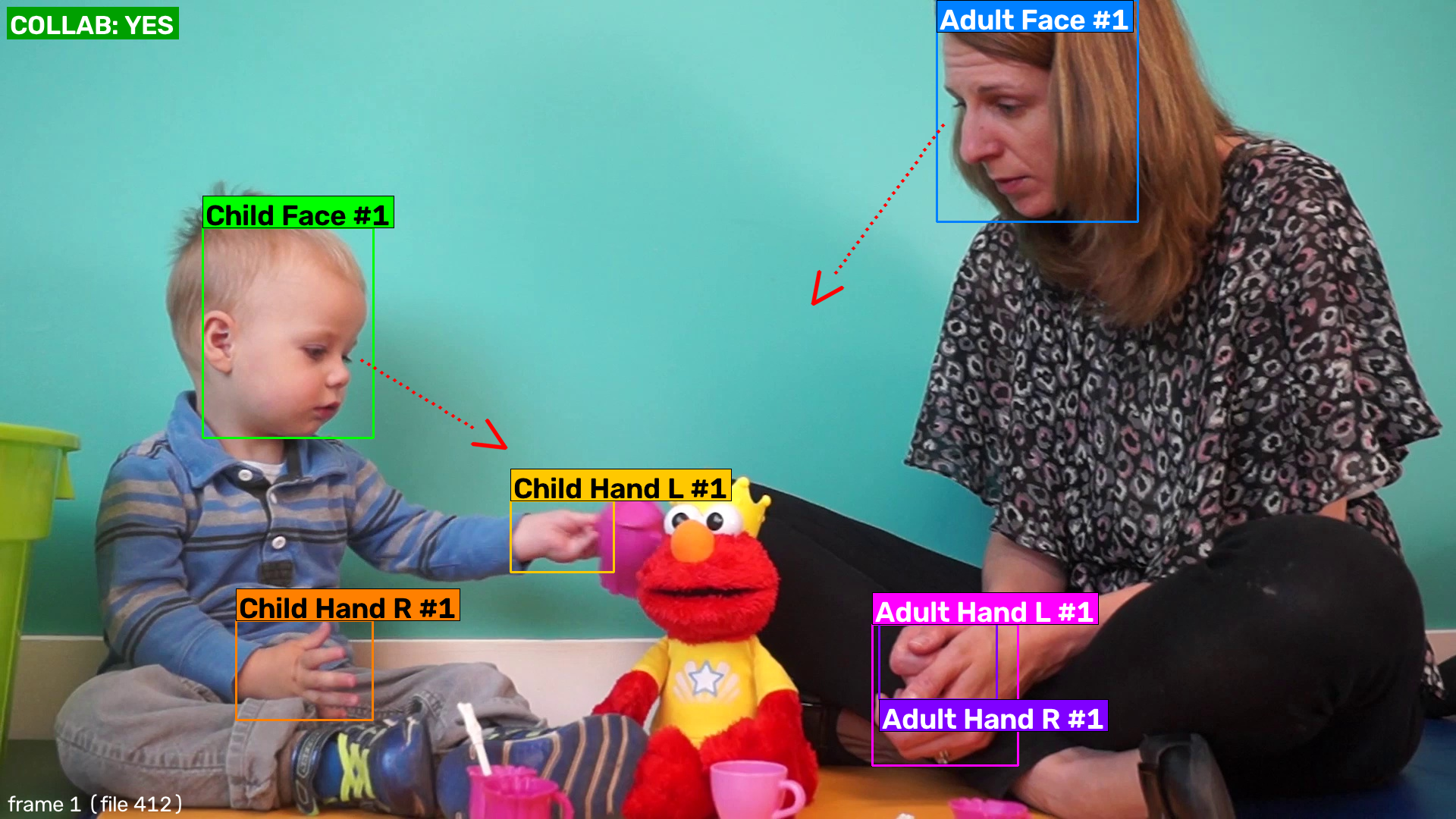}%
    \label{fig:collab-a}}
\hfil
\subfloat[\textbf{Child~\#1 Collaboration: Yes} (looking at the adult);
\textbf{Child~\#2 Collaboration: No} (looking elsewhere, not interacting);
\textbf{Frame Collaboration: Undetermined} (equal Yes/No).]{%
    \includegraphics[width=0.32\textwidth]{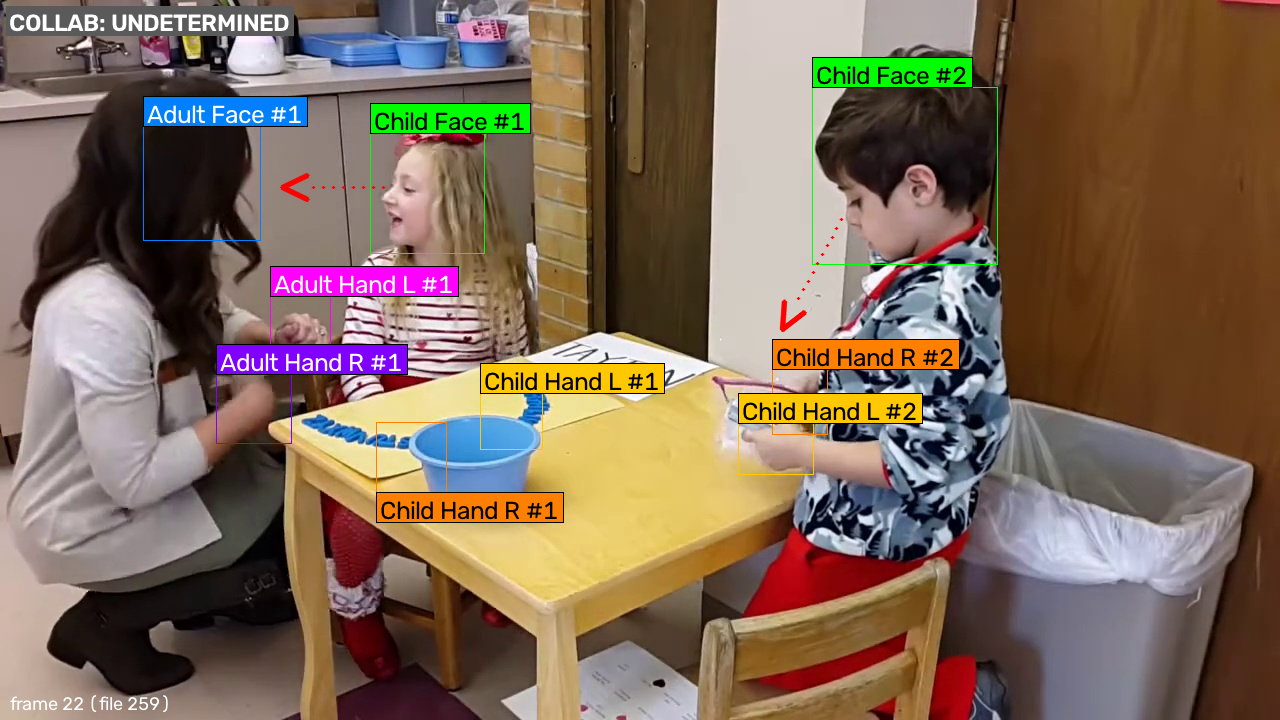}%
    \label{fig:collab-b}}
\hfil
\subfloat[\textbf{Child~\#1 Collaboration: Yes} (looking at the adult);
\textbf{Child~\#2 Collaboration: No} (looking elsewhere, not interacting);
\textbf{Child~\#3 Collaboration: No} (looking elsewhere, not interacting);
\textbf{Frame Collaboration: No} (majority No).]{%
    \includegraphics[width=0.32\textwidth]{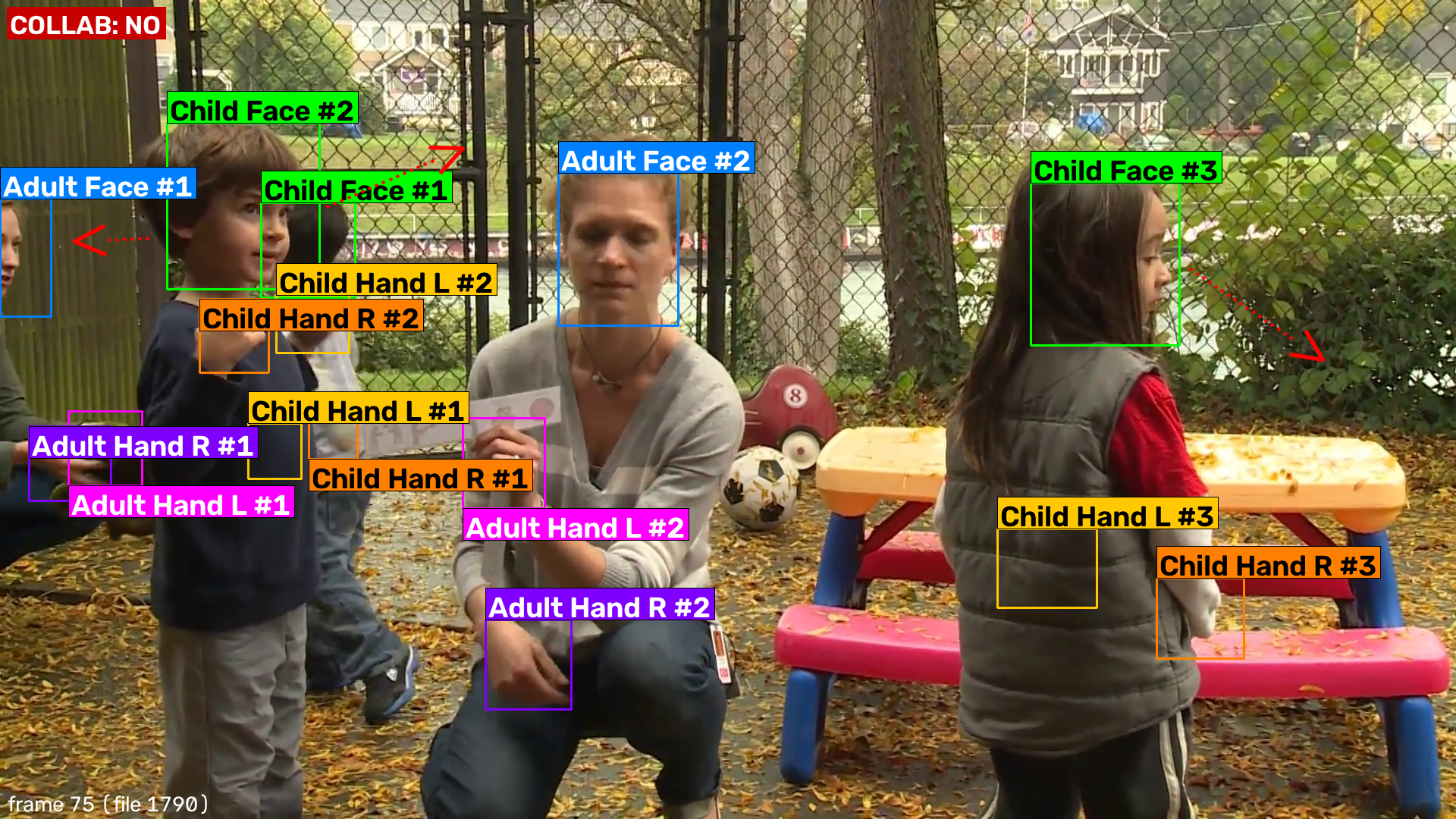}%
    \label{fig:collab-c}}
\\
\subfloat[\textbf{Child~\#1 Collaboration: Yes} (looking at the other child);
\textbf{Child~\#2 Collaboration: No} (looking elsewhere, not interacting);
\textbf{Frame Collaboration: Undetermined} (equal Yes/No).]{%
    \includegraphics[width=0.32\textwidth]{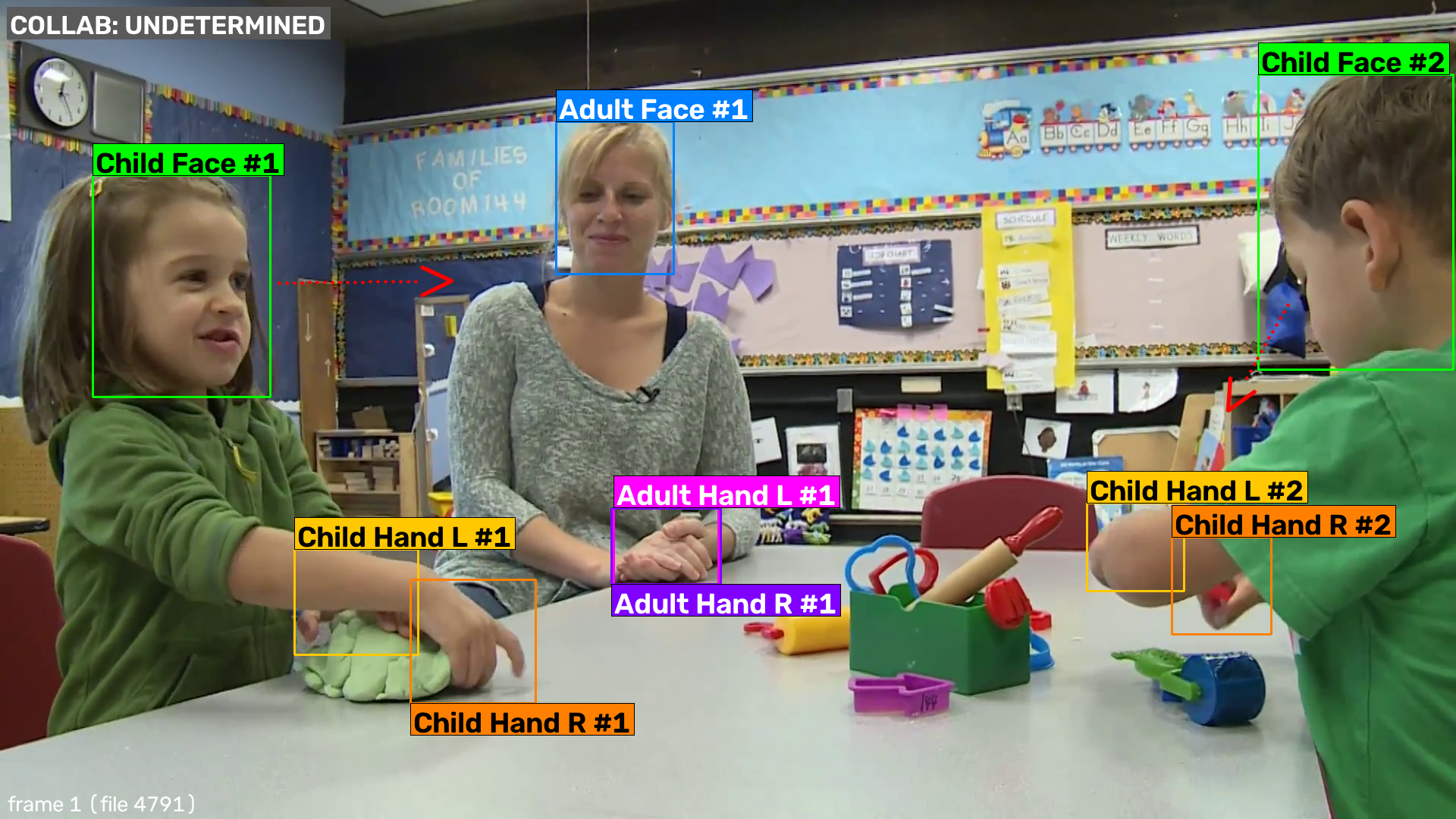}%
    \label{fig:collab-d}}
\hfil
\subfloat[\textbf{Child~\#1 Collaboration: Yes}; \textbf{Child~\#2
Collaboration: Yes}; \textbf{Child~\#3 Collaboration: Yes} (all children
focused on the same object as a team); \textbf{Frame Collaboration: Yes}
(majority Yes).]{%
    \includegraphics[width=0.32\textwidth]{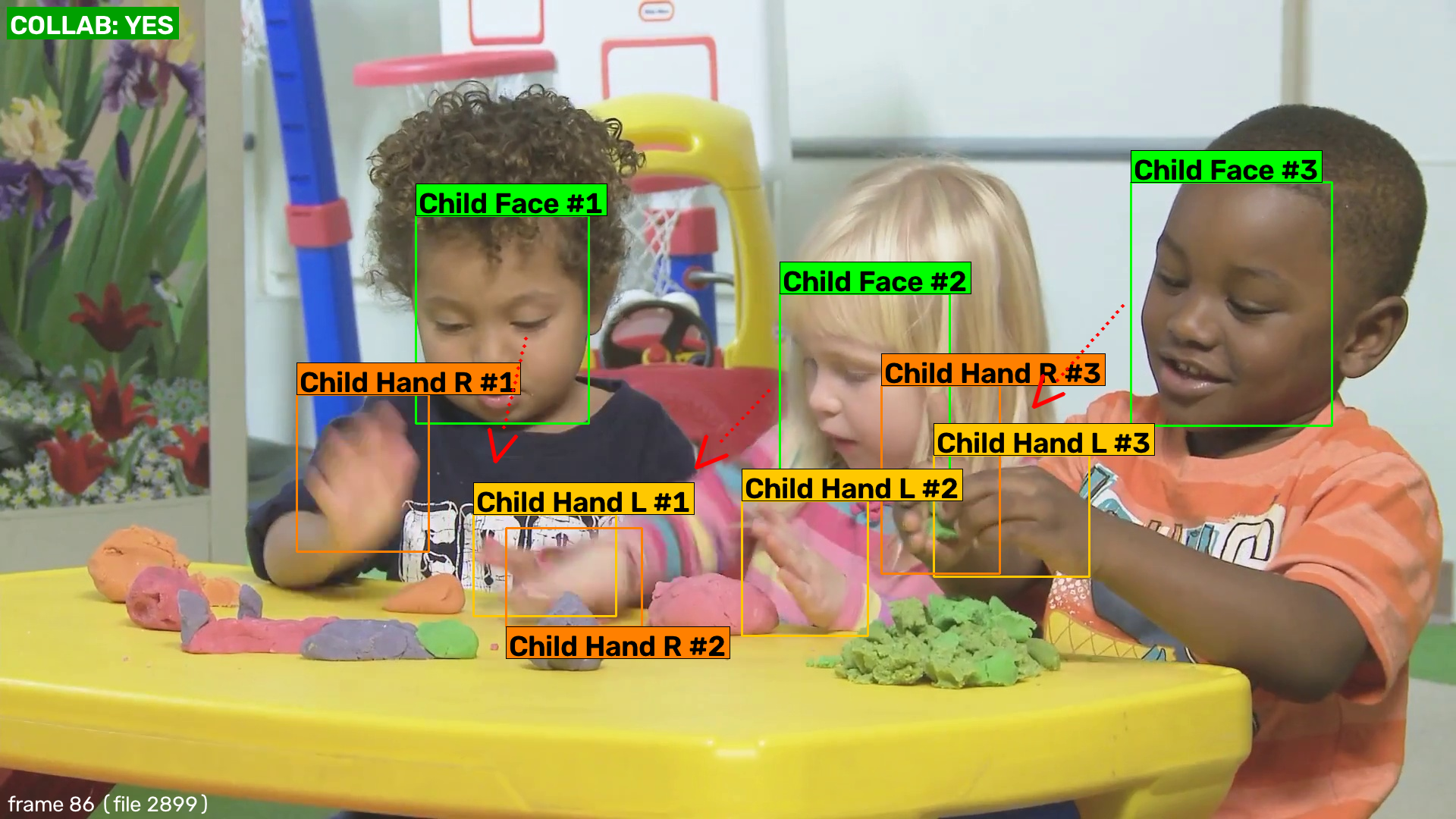}%
    \label{fig:collab-e}}
\hfil
\subfloat[\textbf{Child~\#1 Collaboration: Yes} (child and adult working with
the same object); \textbf{Child~\#2 Collaboration: No} (looking elsewhere, not
interacting); \textbf{Child~\#3 Collaboration: No} (looking elsewhere, not
interacting); \textbf{Frame Collaboration: No} (majority No).]{%
    \includegraphics[width=0.32\textwidth]{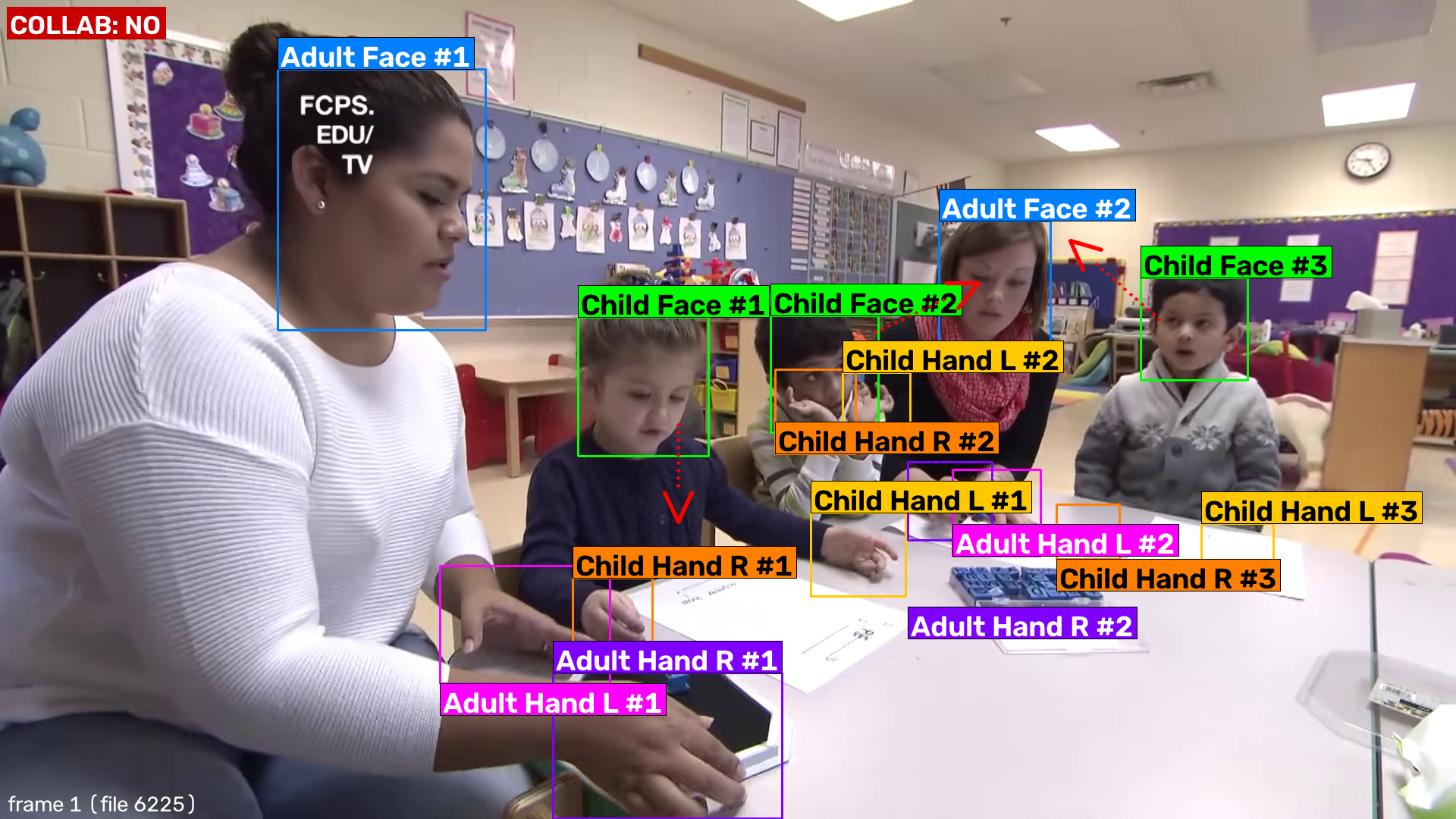}%
    \label{fig:collab-f}}
\\
\subfloat[\textbf{Child~\#1 Collaboration: No} (looking elsewhere, not
interacting); \textbf{Child~\#2 Collaboration: Yes} (interacting with
Child~\#3); \textbf{Child~\#3 Collaboration: Yes} (interacting with
Child~\#2); \textbf{Frame Collaboration: Yes} (majority Yes).]{%
    \includegraphics[width=0.32\textwidth]{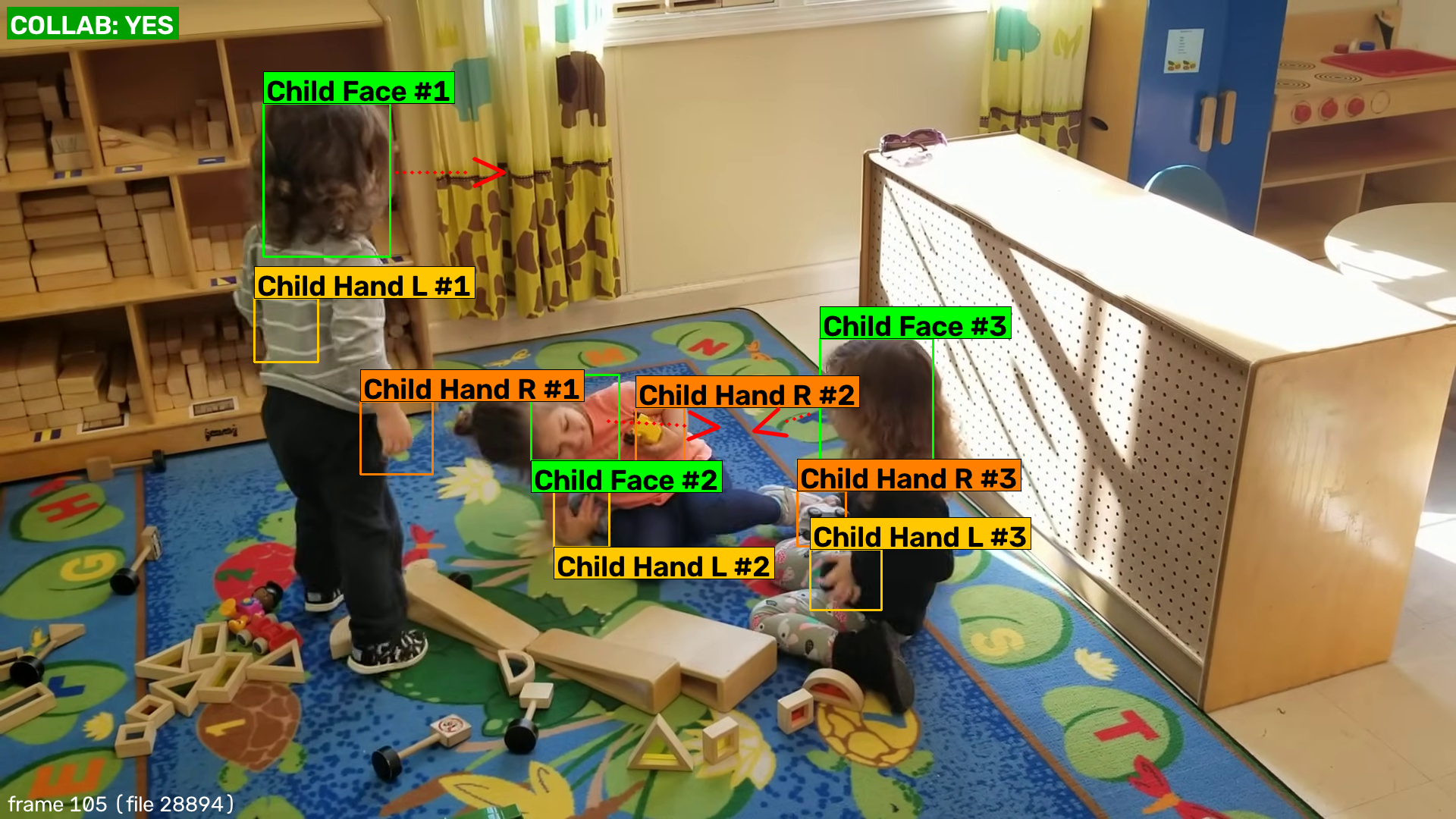}%
    \label{fig:collab-g}}
\hfil
\subfloat[\textbf{Child~\#1 Collaboration: Yes} (child and adult looking at the each other); \textbf{Frame Collaboration: Yes} (majority Yes).]{%
    \includegraphics[width=0.32\textwidth]{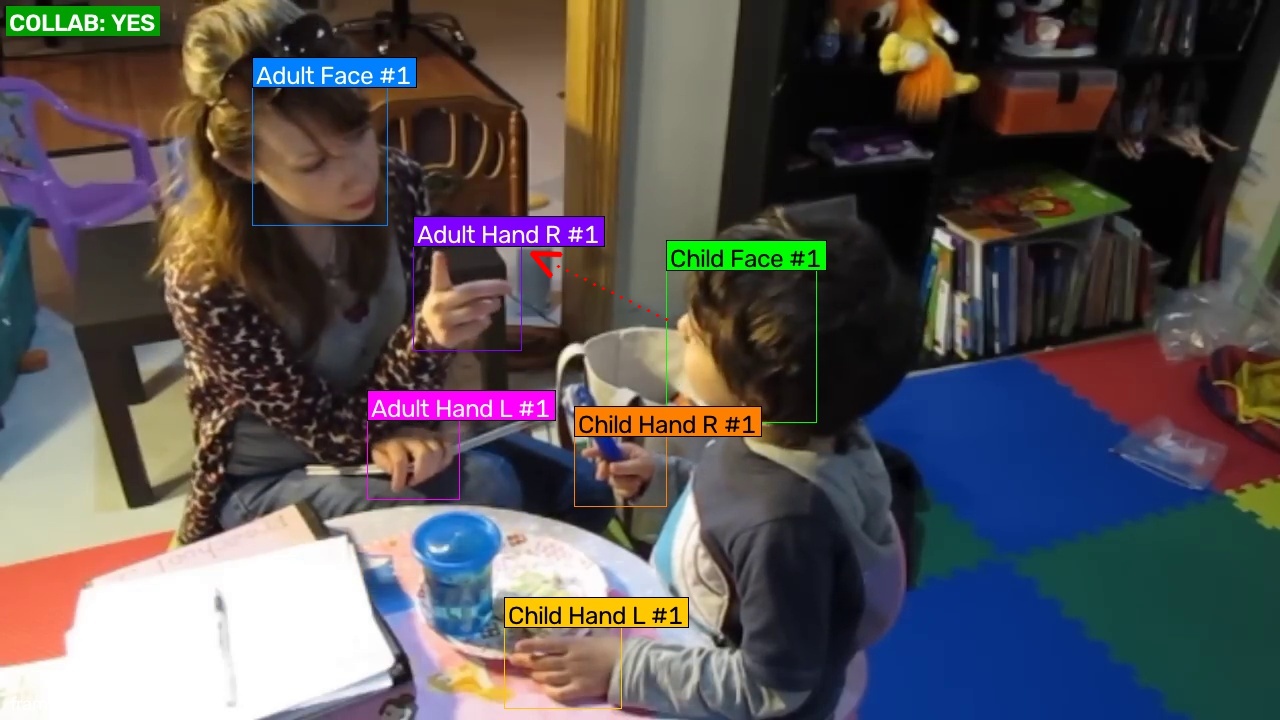}%
    \label{fig:collab-h}}
\hfil
\subfloat[\textbf{Child~\#1 Collaboration: Yes} (child and adult looking at same object); \textbf{Frame Collaboration: Yes} (majority Yes).]{%
    \includegraphics[width=0.32\textwidth]{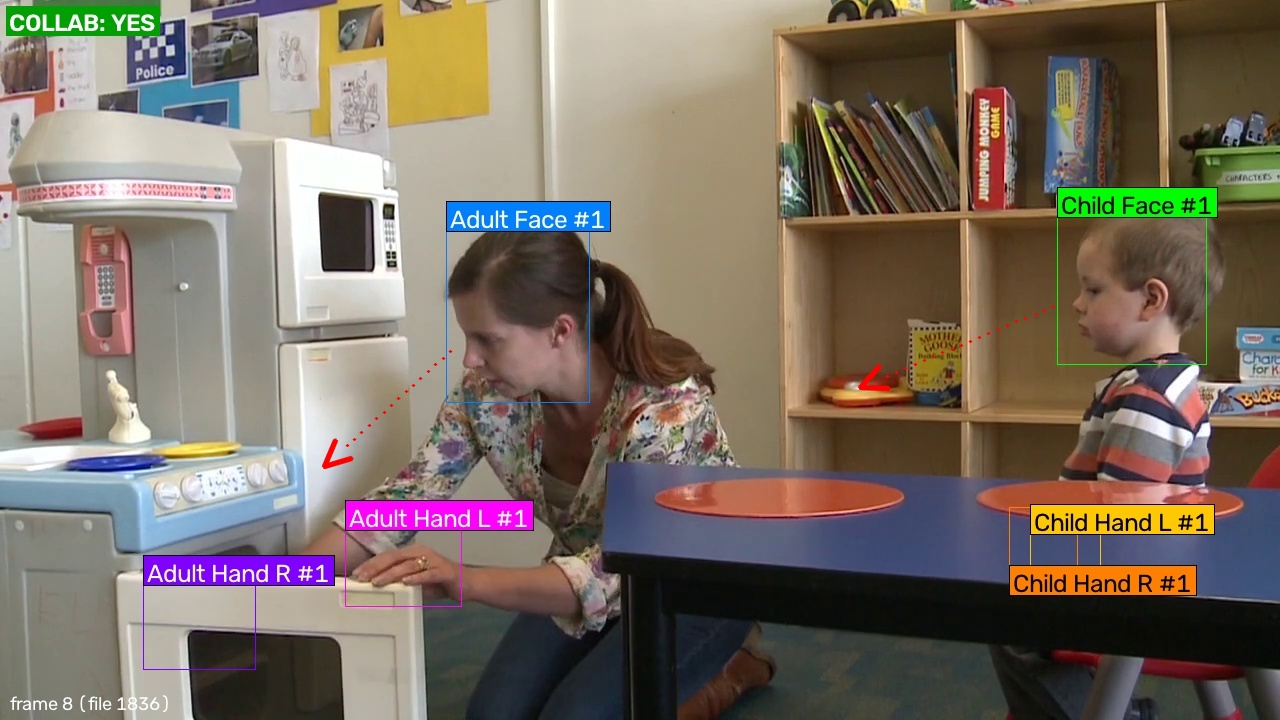}%
    \label{fig:collab-i}}
\caption{Representative examples of the frame-level collaboration annotation
scheme. Each child receives an individual collaboration label, and the frame
label is assigned by majority vote across the children: \textbf{Yes} when the
collaborative labels are in the majority, \textbf{No} when the
non-collaborative labels are in the majority, and \textbf{Undetermined} when
the two counts are equal.}
\label{fig:collaboration-examples}
\end{figure*}

Gaze is an important non-verbal cue for understanding attention and social interaction. In children, gaze provides information about how they attend to people, objects, and ongoing activities. Gaze behaviors such as eye contact and shared attention are important markers of child development and are used in the screening and diagnosis of neuro-developmental disorders such as autism spectrum disorder \cite{tafasca2023childplay}. Early gaze-following research focused on predicting where a person is looking within a visual scene \cite{looking}. However, most public gaze-target benchmarks have been dominated by adult subjects, which can limit their applicability to young children \cite{tafasca2023childplay,cui2017specialized}.

The ChildPlay benchmark addressed this limitation by introducing naturalistic videos of children playing and interacting with adults in uncontrolled environments, including kindergartens, preschools, and therapy settings \cite{tafasca2023childplay}. It provides rich gaze annotations and supports the study of child-centered gaze behavior in realistic interaction settings. More recent approaches, including Sharingan \cite{tafasca2024sharingan} and Gaze-LLE \cite{gazelle}, have further advanced gaze-target estimation using multi-person transformer modeling and pretrained visual representations. ChildPlay-Hand extended the same setting with person, hand, object, and manipulation annotations for studying hand-object interaction \cite{childhand}. Together, these resources improve our ability to analyze where a child is looking and how people interact with objects in naturalistic scenes.

Knowing a child's gaze target, however, does not directly indicate whether the child is participating in a shared activity\cite{jat}. A child may look toward an adult without taking part in the ongoing task. Another child may collaborate with a peer without continuously looking at the adult\cite{asd}. A child may also engage in independent play while other people are present in the scene. These situations require information about the child's participation in the interaction rather than gaze location alone. Recent work using ChildPlay videos has shown that visual cues such as gaze direction, interaction, and peer collaboration can support the recognition of behavioral and collaborative engagement \cite{penchala2025learning}. This motivates the development of annotations that explicitly represent collaborative participation for each child.

In this work, we introduce \emph{ChildGaze}, a new behavioral annotation dataset built from the ChildPlay video collection. We introduce two child-centered behavioral labels, \emph{collaborative} and \emph{non-collaborative}, and assign the collaboration state independently to each child within a frame. A child is considered collaborative when the child participates in a shared interaction, task, or activity with an adult or another child. A child is considered non-collaborative when the child is disengaged from the ongoing interaction or participates in an independent or unrelated activity. Within our annotation scheme, adults are treated as collaborative interaction partners because they typically guide, support, demonstrate, or participate in the activities observed in the selected videos.

The person-wise annotation design is particularly important for scenes containing multiple children. Different children within the same frame may exhibit different collaboration states, and assigning a single label to the entire scene would remove this distinction. Fig. ~\ref{fig:collaboration-examples} illustrates representative examples in which children within the same scene exhibit collaborative and non-collaborative behavior,
as well as cases where the resulting frame-level label is
collaborative, non-collaborative, or undetermined. ChildGaze therefore preserves the behavioral state of each child independently and derives frame-level summaries for broader analysis. By extending child-centered gaze data with explicit collaboration labels, the dataset supports research on child engagement, child-adult interaction, peer interaction, and multi-person behavior understanding. It also provides a foundation for studying how gaze and other visual cues can be combined to recognize collaborative behavior in naturalistic settings.

The main contributions of this work are as follows:
\begin{itemize}
    \item We introduce \emph{ChildGaze}, a behavioral annotation dataset built on the ChildPlay videos with two child-centered labels: \emph{collaborative} and \emph{non-collaborative}.

    \item We provide dense person-wise annotations at the frame level, allowing the collaboration state of each child to be represented independently in multi-person scenes.

    \item We define collaboration based on a child's participation with adults, peers, and shared tasks or activities rather than relying on gaze-target location alone.

    \item We provide a structured annotation and validation framework for assessing the consistency and reliability of the proposed behavioral labels.

    \item We establish ChildGaze as a benchmark for collaborative-behavior recognition and provide baseline experiments using contemporary visual models.
\end{itemize}

The rest of the paper is organized as follows. Section II reviews related work on child-centered behavioral datasets, gaze analysis, and social interaction understanding. Section III presents the construction and annotation process of the ChildGaze dataset, including the annotation protocol and collaboration labeling scheme. Section IV describes the annotation validation procedure and evaluates the reliability and consistency of the annotations, and baseline experiments. Finally, Section V concludes the paper and outlines directions for future work.

\section{Related Work}

This section reviews prior work in gaze following, child-centered behavioral datasets, collaborative engagement recognition, and video annotation reliability, with emphasis on the research most closely related to ChildGaze.

\subsection{Gaze Following and Gaze-Target Estimation}

Gaze following aims to determine where a person is looking within a visual scene. Recasens et al. \cite{looking} formulated gaze following as a visual prediction problem using both head appearance and scene information. Their work established an early benchmark for predicting gaze targets in unconstrained images. Later studies improved gaze prediction through stronger visual representations and more structured reasoning about people and scenes. However, most publicly available gaze-target benchmarks historically contained predominantly adult subjects. Cui et al. \cite{cui2017specialized} specifically studied gaze estimation for children and showed that child gaze estimation benefits from child-specific modeling and domain adaptation.

More recent approaches have improved gaze-target prediction in complex scenes. Sharingan \cite{tafasca2024sharingan} introduces a transformer-based architecture that jointly models multiple people in an image and predicts their gaze targets in a single forward pass. Gaze-LLE \cite{gazelle} follows a different approach by using features from a frozen pretrained DINOv2 encoder together with a lightweight person-specific gaze decoder. These methods achieve strong gaze-target estimation performance, but their primary objective remains determining where a person is looking. They do not explicitly represent whether a child is participating in a shared activity or interaction.

\subsection{Child-Centered Gaze and Behavior Datasets}

ChildPlay \cite{tafasca2023childplay} is a child-centered benchmark designed to study gaze behavior in naturalistic environments. It contains short video clips of children playing and interacting with adults in uncontrolled settings such as kindergartens, preschools, and therapy centers. The dataset provides rich gaze annotations and was introduced in response to the limited representation of children in existing public gaze-target benchmarks. ChildPlay also demonstrated that gaze prediction performance can differ substantially between adults and children and that child-specific annotations can improve performance.

Subsequent datasets have extended child-centered visual analysis beyond gaze localization. ChildPlay-Hand \cite{childhand} adds person and object bounding boxes, per-hand annotations, and manipulation actions in naturalistic child-adult interactions. It also retains gaze information from ChildPlay, enabling joint analysis of gaze and hand-object interaction. MMASD \cite{mmasd} focuses on privacy-preserving behavior analysis for children with Autism Spectrum Disorder and provides multimodal representations including 3D skeletons, 3D body meshes, and optical flow. More recently, the Autism Gaze Target dataset \cite{agt} introduced gaze-target annotations for young autistic children observed during naturalistic play sessions. These resources demonstrate the growing need for child-specific datasets that capture social attention, actions, and interaction context.

\subsection{Engagement and Collaborative Behavior Recognition}

Visual analysis of student and child behavior has also been studied in educational settings. SCB-Dataset3 \cite{scb}, for example, contains classroom images annotated with behaviors such as hand raising, reading, writing, phone use, bowing the head, and leaning over a table. The dataset supports detection of observable classroom behaviors across realistic educational environments. However, these labels primarily describe individual actions rather than whether a child is participating collaboratively with another person.

Collaborative engagement requires reasoning about relationships among people and their participation in a shared activity. Penchala et al. \cite{penchala2025learning} investigated behavioral and collaborative engagement using ChildPlay videos and Vision Transformer architectures. Their study used visual cues such as gaze direction, interaction, and peer collaboration to classify engagement states. The results showed that visual representations can support automatic recognition of collaborative engagement. In contrast to that classification-oriented study, ChildGaze focuses on constructing and validating person-wise collaboration annotations as a reusable benchmark resource.

\begin{table*}[t]
\caption{Comparison of ChildGaze with existing child-centered gaze, behavior, and interaction datasets in terms of annotation type, spatial cues, multi-person support, behavioral labeling, annotation granularity, and evaluation methodology.}
\label{tab:dataset_comparison}
\centering
\renewcommand{\arraystretch}{1.3}
\setlength{\tabcolsep}{2pt}
\small

\begin{tabular}{lcccccc}
\hline
\textbf{Feature} &
\textbf{ChildPlay\cite{tafasca2023childplay}} &
\textbf{ChildPlay-Hand\cite{childhand}} &
\textbf{ChildPlay-R\cite{cpr}} &
\textbf{SCB-Dataset3\cite{scb}} &
\textbf{MMASD\cite{mmasd}} &
\textbf{ChildGaze (Ours)} \\
\hline

Multiple children per scene
& \cmark & \cmark & \xmark & \cmark & \xmark & \textbf{\cmark} \\


Face/head bounding boxes
& \cmark & \xmark & \cmark & \xmark & \xmark & \textbf{\cmark} \\

Hand bounding boxes
& \xmark & \cmark & \xmark & \xmark & \xmark & \textbf{\cmark} \\


Action/behavior labels
& \xmark & \cmark & \xmark & \cmark & \cmark & \cmark \\

Person-level annotation
& \cmark & \cmark & \cmark & \cmark & \cmark & \textbf{\cmark} \\

Frame-level annotation
& \cmark & \cmark & \cmark & \cmark & \xmark & \textbf{\cmark} \\

Row-level annotation
& \cmark & \cmark & \xmark & \xmark & \xmark & \textbf{\cmark} \\

\hline

\textbf{Evaluation metrics}
& \makecell{AUC, Dist.,\\ AP, P.Head}
& \makecell{Acc., P, R, F1,\\ S-F1, Edit}
& \makecell{AUC, out-of-\\sample Prec.}
& \makecell{mAP@50, BSI\\ mAP@50:95, P, R}
& \makecell{Action\\ accuracy}
& \makecell{\textbf{Cohen's $\kappa$}, \textbf{IoU}, \\Acc. , Pre, \\Recall, F1, AUC} \\

\hline
\end{tabular}

\vspace{2pt}


\end{table*}

\subsection{Video Annotation and Dataset Validation}

The usefulness of a behavioral dataset depends not only on its scale but also on the consistency of its annotations. This is particularly important for collaboration labels because they represent a behavioral interpretation of social participation rather than a directly observable object category. Clear labeling criteria and independent annotation checks are therefore necessary to quantify reliability.

Recent dataset research has also emphasized the cost and quality of video annotation. FOCAL \cite{focal} was introduced as a cost-aware video dataset that records real annotation time and quality-assurance effort for video sequences. Its design highlights the importance of documenting the annotation process rather than treating annotation as an unreported preprocessing step. For categorical behavioral labels, Cohen's kappa \cite{ckc} provides a chance-corrected measure of agreement between two annotators. Landis and Koch \cite{landis} provide commonly used qualitative ranges for interpreting kappa values. Following this general reliability framework, ChildGaze uses independent annotation of a held-out subset to evaluate both collaboration-label agreement and bounding-box consistency.

To summarize the differences among the most relevant child-centered datasets, Table~\ref{tab:dataset_comparison} compares their annotation types, spatial cues, behavioral labels, annotation granularity, and evaluation strategies. Existing datasets provide important resources for gaze estimation, hand-object interaction, classroom behavior, and autism-related analysis. However, these datasets generally focus on a single aspect of behavior or interaction. In contrast, ChildGaze combines gaze context with face and hand annotations and introduces explicit collaboration labels at multiple levels of analysis.

As shown in Table~\ref{tab:dataset_comparison}, ChildGaze differs from prior datasets in the combination of features it provides. ChildPlay and ChildPlay-R\cite{cpr} focus mainly on gaze behavior, while ChildPlay-Hand extends this setting with hand and object annotations. SCB-Dataset3 and MMASD emphasize behavioral or activity-related analysis but do not provide collaboration labels at the individual-child level. ChildGaze brings these directions together by supporting gaze information, face and hand annotations, multi-person scenes, and explicit collaborative versus non-collaborative labels. The annotations are further organized at the row, person, and frame levels and are validated using both Cohen's $\kappa$ and IoU-based spatial agreement.

\section{ChildGaze Dataset Construction and Annotation}

\begin{figure*}[tp]
    \centering
    \includegraphics[width=0.95\textwidth]{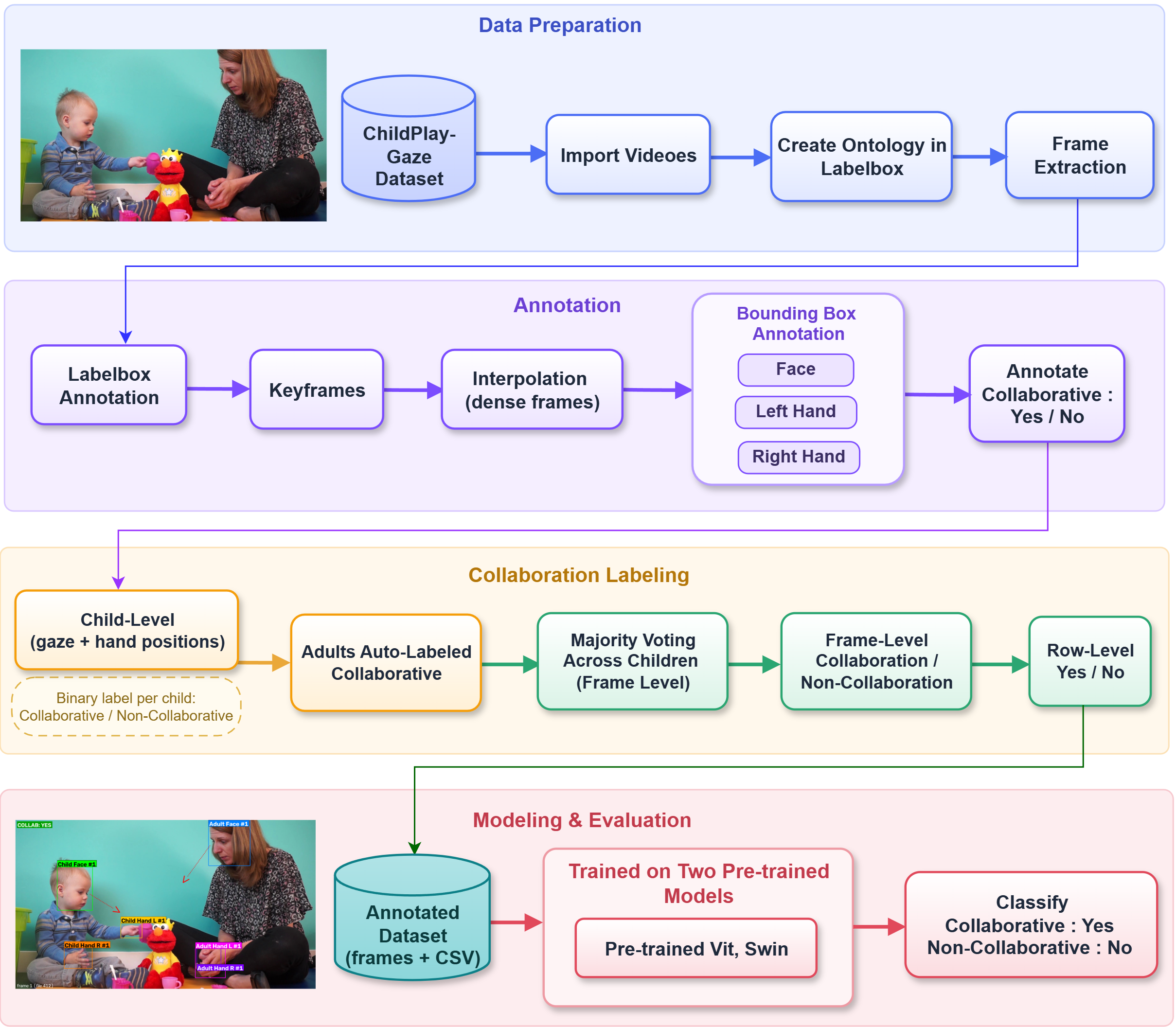}
    \caption{Overview of the ChildGaze dataset construction and analysis pipeline. Source ChildPlay \cite{tafasca2023childplay}  videos are prepared and imported into Labelbox, where separate child and adult annotation classes are defined. Face, left-hand, and right-hand bounding boxes are annotated and propagated across frames, followed by assignment of collaborative or non-collaborative labels to individual children. The resulting annotations are organized at the row, person, and frame levels to preserve both individual and scene-level interaction information. The final annotated dataset supports validation and downstream evaluation of collaborative-behavior recognition using pretrained visual models\cite{tafasca2023childplay}.}    
    \label{fig:childgaze_pipeline}
\end{figure*}

This section describes the construction of the ChildGaze dataset and the annotation scheme used to represent collaborative behavior. ChildGaze is developed from videos in the ChildPlay benchmark \cite{tafasca2023childplay}. The source videos contain naturalistic interactions among children, adults, and peers during play and task-oriented activities. We extend these videos with person-wise collaboration labels and dense frame-level annotations for children and adults. Fig. ~\ref{fig:childgaze_pipeline} summarizes the complete dataset preparation, annotation, collaboration labeling, and evaluation pipeline.

As illustrated in Fig. ~\ref{fig:childgaze_pipeline}, the workflow consists of four main stages. First, the source videos are prepared and imported into Labelbox. Second, faces and hands are annotated for children and adults using bounding boxes. Third, collaboration labels are assigned at the child-person level and summarized at the frame level. Finally, the structured annotations are exported for validation and downstream collaborative-behavior modeling.


\subsection{Dataset Preparation}

The ChildPlay videos were imported into Labelbox for manual annotation. Separate annotation classes were defined for children and adults. For each visible person, annotators identified the face, left hand, and right hand using bounding boxes. Instance identifiers were used to distinguish different people when multiple children or adults appeared in the same frame.

Annotations were created on video frames using the tracking and interpolation tools available in Labelbox. Bounding boxes were placed on keyframes and propagated across intermediate frames. Annotators reviewed the propagated annotations and corrected them when required. The completed annotations were exported in NDJSON format and converted into individual CSV files. Each CSV record contains information such as the video identifier, frame number, person category, body-part label, instance identifier, bounding-box coordinates, and collaboration label. Together, the final dataset contains 27 annotated files and 10,641 frames.

\subsection{Collaboration Annotation Scheme}

The primary behavioral annotation in ChildGaze represents the collaboration state of each child. Two labels are used: \emph{collaborative} and \emph{non-collaborative}. Labels are assigned independently to each child within a frame. The decision is based on the child's participation in the ongoing interaction, including attention to other participants and involvement in a shared activity or task.

A child is labeled \emph{collaborative} when the child participates in a shared interaction with an adult or another child. This includes attending to an adult during an activity, performing a task together with an adult, or working with peers on the same object or activity. Continuous visual attention toward the adult is not required when the child is clearly participating in the shared task. A child is labeled \emph{non-collaborative} when the child is disengaged from the ongoing interaction, directs attention toward an unrelated activity, or plays independently without shared participation.

Adults are treated differently from children in the annotation scheme. In the selected ChildPlay interactions, adults generally guide, demonstrate, encourage, assist, or participate in activities with the children. Therefore, adult instances are recorded as collaborative interaction partners. The collaborative versus non-collaborative behavioral decision is primarily defined for children. Table~\ref{tab:collaboration_rules} summarizes the main rules used for assigning child collaboration labels.

\begin{table}[!t]
\caption{Collaboration Labeling Criteria Used for Child Annotations in ChildGaze}
\label{tab:collaboration_rules}
\centering
\renewcommand{\arraystretch}{1.2}
\begin{tabular}{@{}>{\centering\arraybackslash}p{0.22\columnwidth}p{0.68\columnwidth}@{}}
\hline
\textbf{Label} & \textbf{Annotation Criteria} \\
\hline
\multirow{1}{0.22\columnwidth}[-5.5em]{\centering\textbf{Collaborative}} &
\begin{itemize}[\IEEEsetlabelwidth{$\bullet$}%
\setlength{\IEEElabelindent}{0pt}\setlength{\IEEEiedtopsep}{0pt}%
\setlength{\topsep}{0pt}\setlength{\itemsep}{1pt}]
\item Child listens to an adult and participates in the ongoing interaction.
\item Performs, attempts, or continues a shared task with an adult.
\item Plays with an object that is part of a shared activity.
\item Works or plays with another child on the same task or object.
\item Looks away but remains actively involved in the shared activity.
\end{itemize} \\[3pt]
\hline
\multirow{1}{0.22\columnwidth}[-3.5em]{\centering\textbf{Non-Collaborative}} &
\begin{itemize}[\IEEEsetlabelwidth{$\bullet$}%
\setlength{\IEEElabelindent}{0pt}\setlength{\IEEEiedtopsep}{0pt}%
\setlength{\topsep}{0pt}\setlength{\itemsep}{1pt}]
\item Child does not pay attention to the ongoing interaction and does not participate.
\item Plays or works independently without shared participation.
\item Attends to an unrelated person, object, or activity.
\item Is present in the scene but shows no observable participation in the shared interaction.
\end{itemize} \\[3pt]
\hline
\end{tabular}
\end{table}

\subsection{Annotation Levels}

ChildGaze annotations are organized at three levels: row level, person level, and frame level. Each level represents a different unit of analysis.

\textit{Row level:}
Each annotated body part is stored as a separate record. A person may therefore contribute up to three rows in a frame, corresponding to the face, left hand, and right hand. If a body part is not visible or cannot be reliably localized, the corresponding annotation is omitted.

\textit{Person level:}
Each child is counted once per frame regardless of the number of available body-part annotations. The collaboration label is assigned at this level and represents the primary unit for behavioral analysis. This allows children appearing in the same frame to receive different collaboration labels.

\textit{Frame level:}
A frame-level label summarizes the collaboration states of all children visible in a frame. A frame is labeled collaborative when the majority of children are collaborative and non-collaborative when the majority are non-collaborative. Frames with an equal number of collaborative and non-collaborative children are marked as \emph{undetermined}. The frame-level representation is therefore derived from the underlying person-level annotations.

\subsection{Dataset Statistics}

The final ChildGaze dataset contains 27 annotated files and 10,641 frames collected in two annotation phases. In total, the dataset contains 73,268 body-part annotation rows across children and adults.

At the person level, 13,330 child instances are annotated. Among them, 5,959 (44.7\%) are labeled collaborative and 7,371 (55.3\%) are labeled non-collaborative. At the frame level, 4,258 frames are collaborative, 5,730 are non-collaborative, and 653 are undetermined. The undetermined category represents frames in which collaborative and non-collaborative children occur in equal numbers.

The body-part annotations include faces and both hands for children and adults whenever they are visible. Small differences among face and hand counts occur because hands may be occluded, outside the image boundary, or otherwise unsuitable for reliable annotation. Adults appear in most of the annotated frames and provide contextual information about the ongoing interaction. Their collaboration labels are not used as targets for the child collaboration classification task.







\begin{table}[tp]
\caption{Detailed statistics of the ChildGaze dataset across frame-, person-, and body-part-level annotations.}
\label{tab:dataset_statistics}
\centering
\begin{tabular}{lrrr}
\hline
\textbf{Property} & \textbf{Yes} & \textbf{No} & \textbf{Total} \\
\hline

\multicolumn{4}{l}{\textbf{\textit{Dataset Overview}}} \\
Annotated files & -- & -- & 27 \\
Total frames & -- & -- & 10,641 \\
Total annotation rows & 51,155 & 22,113 & 73,268 \\
\hline

\multicolumn{4}{l}{\textbf{\textit{Child Annotations}}} \\
Child annotation rows & 17,654 & 22,113 & 39,767 \\
Child person-instances & 5,959 & 7,371 & 13,330 \\
Child frames$^{a}$ & 4,258 & 5,730 & 10,641 \\
\hline

\multicolumn{4}{l}{\textbf{\textit{Child Body-Part Annotations}}} \\
Child face annotations & 5,959 & 7,371 & 13,330 \\
Child left-hand annotations & 5,812 & 7,371 & 13,183 \\
Child right-hand annotations & 5,883 & 7,371 & 13,254 \\
\hline

\multicolumn{4}{l}{\textbf{\textit{Adult Annotations}}} \\
Adult annotation rows & 33,501 & 0 & 33,501 \\
Adult person-instances & 11,258 & 0 & 11,258 \\
\hline

\multicolumn{4}{l}{\textbf{\textit{Adult Body-Part Annotations}}} \\
Adult face annotations & 11,275 & 0 & 11,275 \\
Adult left-hand annotations & 11,234 & 0 & 11,234 \\
Adult right-hand annotations & 10,992 & 0 & 10,992 \\
\hline

\multicolumn{4}{l}{\textbf{\textit{Adult Frame Coverage}}} \\
Frames with adult annotations & -- & -- & 10,272 \\
Frames without adult annotations & -- & -- & 369 \\
\hline
\multicolumn{4}{p{0.94\columnwidth}}{\footnotesize
$^{a}$ In addition to the reported Yes and No frame labels, 653 frames were marked as undetermined because collaborative and non-collaborative children were equally represented.} \\
\end{tabular}
\end{table}

Table~\ref{tab:dataset_statistics} provides the complete distribution of annotations at the row, person, frame, and body-part levels. The person-level distribution is relatively balanced, with a modest prevalence of non-collaborative behavior. This is useful for benchmark development because both collaboration states are represented at substantial scale without artificially balancing the natural behavior distribution. The difference between the number of frames and child person-instances also reflects the presence of multi-child interactions, where collaboration must be determined separately for each child. These properties make the person-level annotations the most appropriate representation for the primary ChildGaze benchmark.

\section{Annotation Validation and Reliability}

To assess the reliability of the ChildGaze annotations, a subset of the dataset was independently annotated by two annotators. Validation was performed separately for the child collaboration labels and the spatial bounding-box annotations. Agreement on the collaborative or non-collaborative labels was measured using Cohen's kappa, while bounding-box consistency was evaluated using Intersection over Union (IoU), agreement at an IoU threshold of 0.5, and the missed-instance rate \cite{ckappa,iou}.

\subsection{Validation Subset}

The validation subset contains 1,187 frames selected from three annotated video segments. This represents approximately 11.2\% of the 10,641 frames in the complete dataset. Both annotators labeled the same frames using the same annotation ontology and collaboration criteria. Each annotator produced 8,508 annotation rows, including 4,254 child annotations and 4,254 adult annotations. After grouping the child face and hand annotations by person and frame, the subset contained 1,418 child person-instances. These shared person-instances were used to evaluate agreement on the collaborative behavior labels.

\subsection{Agreement on Collaboration Labels}

Cohen's kappa ($\kappa$) \cite{ckinter} was used to measure inter-annotator agreement for the binary child collaboration labels. Unlike raw agreement, Cohen's kappa accounts for agreement that may occur by chance. It is defined as

\begin{equation}
\kappa = \frac{P_o-P_e}{1-P_e},
\end{equation}

where, $P_o$ is the observed agreement between annotators and $P_e$ is the expected agreement by chance based on the marginal label distributions.

Across the 1,418 child person-instances, the two annotators achieved a raw agreement of 93.16\%. Cohen's kappa was 0.8631. According to the interpretation proposed by Landis and Koch \cite{landis}, this value falls within the \emph{almost perfect} agreement range. Table~\ref{tab:kappa_validation} summarizes the agreement between the two annotators. Of the 1,418 instances, both annotators labeled 644 as collaborative and 677 as non-collaborative. Disagreement occurred in 97 cases. Annotator A labeled 59 instances as collaborative that Annotator B labeled as non-collaborative, while 38 instances showed the opposite disagreement.

\begin{table}[t]
\caption{Inter-annotator agreement for child collaboration labels in the validation subset\cite{ckinter}.}
\label{tab:kappa_validation}
\centering
\begin{tabular}{lcc}
\hline
 & \textbf{Annotator B: Yes} & \textbf{Annotator B: No} \\
\hline
\textbf{Annotator A: Yes} & 644 & 59 \\
\textbf{Annotator A: No}  & 38  & 677 \\
\hline
\end{tabular}

\vspace{2mm}
\begin{tabular}{lr}
\hline
\textbf{Metric} & \textbf{Value} \\
\hline
Compared person-instances & 1,418 \\
Raw agreement & 93.16\% \\
Cohen's $\kappa$ & 0.8631 \\
Annotator A Yes prevalence & 49.6\% \\
Annotator B Yes prevalence & 48.1\% \\
\hline
\end{tabular}
\end{table}

The similar prevalence of collaborative labels for the two annotators also indicates that the agreement was not produced by one annotator consistently favoring one class. Annotator A assigned the collaborative label to 49.6\% of the validation instances, compared with 48.1\% for Annotator B.

\subsection{Bounding-Box Agreement}

Spatial consistency was evaluated independently for the six body-part classes: child face, child left hand, child right hand, adult face, adult left hand, and adult right hand. Bounding boxes from the two annotators were matched within each frame and class using their spatial overlap\cite{iou}. For each matched pair, IoU was calculated as

\begin{equation}
\mathrm{IoU}(B_A,B_B)=
\frac{|B_A \cap B_B|}
{|B_A \cup B_B|},
\end{equation}

where, $B_A$ and $B_B$ denote bounding boxes produced by Annotators A and B, respectively. An IoU of 1 indicates identical boxes, while an IoU of 0 indicates no spatial overlap.

In addition to mean IoU, we report the proportion of matched annotations reaching an IoU of at least 0.5. This threshold provides a direct measure of whether the two annotations have sufficient spatial agreement. We also report the missed-instance rate to identify cases in which an annotated body part from one annotator could not be matched with a corresponding annotation from the other. Table~\ref{tab:iou_validation} presents the per-class results.

\begin{table}[t]
\caption{Per-class inter-annotator agreement for bounding-box annotations\cite{iou}.}
\label{tab:iou_validation}
\centering
\begin{tabular}{lccc}
\hline
\textbf{Class} &
\textbf{Mean IoU} &
\textbf{IoU $\geq$ 0.5} &
\textbf{Missed instance rate} \\
\hline
Child Face   & 0.892 & 99.9\% & 0.0\% \\
Child Hand L & 0.661 & 65.1\% & 0.1\% \\
Child Hand R & 0.689 & 62.8\% & 0.0\% \\
Adult Face   & 0.896 & 98.6\% & 0.0\% \\
Adult Hand L & 0.839 & 90.4\% & 1.1\% \\
Adult Hand R & 0.871 & 93.7\% & 0.0\% \\
\hline
Overall      & 0.808 & 85.1\% & 0.2\% \\
\hline
\end{tabular}
\end{table}









Face annotations show the strongest spatial consistency. The mean IoU is 0.892 for child faces and 0.896 for adult faces, with nearly all matched face annotations exceeding the 0.5 IoU threshold. Adult hand annotations also show high agreement, with mean IoU values of 0.839 and 0.871 for the left and right hands, respectively.

\begin{table*}[t]

\caption{Baseline performance and inference efficiency of ViT and Swin Transformer for collaborative-behavior recognition at the frame and child-person levels.}
\label{tab:baseline_results}
\centering
\renewcommand{\arraystretch}{1.15}
\setlength{\tabcolsep}{3.5pt}

\begin{tabular}{llcccccccc}
\hline
\textbf{Model} &
\textbf{Level} &
\textbf{Accuracy} &
\textbf{Precision} &
\textbf{Recall} &
\textbf{F1} &
\textbf{Macro-F1} &
\textbf{ROC-AUC} &
\textbf{Latency} &
\textbf{FPS} \\
&
&
&
&
&
&
&
&
\textbf{(ms/frame)} &
\\
\hline

\multirow{2}{*}{\textbf{ViT}}
& Frame
& 0.9640
& \textbf{0.9516}
& 0.9687
& 0.9601
& 0.9636
& 0.9929
& \textbf{1.64}
& \textbf{609.5} \\

&
Child-Person
& \textbf{0.9744}
& \textbf{1.0000}
& 0.9412
& \textbf{0.9697}
& \textbf{0.9737}
& \textbf{1.0000}
& \textbf{1.64}
& \textbf{609.5} \\

\hline

\multirow{2}{*}{\textbf{Swin}}
& Frame
& \textbf{0.9680}
& 0.9501
& 0.9799
& \textbf{0.9648}
& \textbf{0.9677}
& \textbf{0.9957}
& 1.94
& 515.2 \\

&
Child-Person
& 0.9487
& 0.8947
& \textbf{1.0000}
& 0.9444
& 0.9484
& 0.9893
& 1.94
& 515.2 \\

\hline
\end{tabular}
\end{table*}

Child hand annotations show lower spatial agreement, with mean IoU values of 0.661 for the left hand and 0.689 for the right hand. This difference is expected because hands occupy smaller image regions and may undergo rapid motion, partial occlusion, or interaction with objects. Small differences in box placement therefore have a greater effect on IoU for hands than for larger and more stable face regions.

Across all six annotation classes, 8,492 box pairs were successfully matched. The pooled mean IoU was 0.808, and 85.1\% of the matched annotations reached an IoU of at least 0.5. The overall missed-instance rate was only 0.2\%. These results indicate that disagreements were primarily related to the precise localization of small body regions rather than to whether the annotated person or body part was present.



\subsection{Baseline Experiments}

To provide an initial benchmark for collaborative-behavior recognition, we evaluated two pretrained transformer architectures: Vision Transformer (ViT) and Swin Transformer\cite{vit, swin}. Both models were fine-tuned to distinguish between the \emph{collaborative} and \emph{non-collaborative} classes. Performance was evaluated at both the frame level and the child-person level using accuracy, precision, recall, F1-score, macro-F1, and area under the ROC curve (ROC-AUC). We also report inference latency to provide an indication of computational efficiency. Table~\ref{tab:baseline_results} summarizes the results. At the frame level, both models achieved strong performance, with Swin obtaining the highest accuracy of 96.80\%, F1-score of 96.48\%, and ROC-AUC of 0.9957. ViT achieved a comparable accuracy of 96.40\% and ROC-AUC of 0.9929. At the child-person level, ViT achieved the highest overall accuracy of 97.44\%, an F1-score of 96.97\%, and a ROC-AUC of 1.000. Swin obtained lower person-level accuracy of 94.87\%, but achieved a recall of 100\%, indicating that all collaborative child instances in the test set were identified.

The two models also showed low inference latency. ViT required approximately 1.64~ms per frame, corresponding to 609.5 frames/s, while Swin required 1.94~ms per frame, corresponding to 515.2 frames/s, using a batch size of 32 on GPU. These results provide an initial indication that the proposed collaboration labels contain learnable visual patterns and can support both frame-level and person-level collaborative-behavior recognition.

\subsection{Discussion}

The results indicate that collaborative behavior can be annotated consistently at the individual-child level in naturalistic ChildPlay scenes. The person-wise design is important because children appearing in the same frame may show different levels of participation, which cannot always be represented by a single frame-level label. This is also reflected in the 653 undetermined frames, where collaborative and non-collaborative children were equally represented. Compared with existing resources such as ChildPlay \cite{tafasca2023childplay} and ChildPlay-Hand \cite{childhand}, which mainly describe gaze and hand-object interactions, ChildGaze adds a higher-level behavioral interpretation of whether a child is participating in a shared activity. The validation results further support the consistency of this annotation scheme, with 93.16\% raw agreement and a Cohen's kappa of 0.8631 for collaboration labels.

The current validation was performed on 1,187 frames, representing approximately 11.2\% of the complete dataset. Although this subset provides an initial measure of annotation reliability, validating a larger portion of the dataset would provide stronger evidence of consistency across different videos, activities, and interaction settings. The lower IoU observed for child hands also suggests that small and frequently occluded body regions remain more difficult to annotate consistently than faces. Future validation will therefore include additional samples and a broader range of interaction scenarios. Overall, the current findings suggest that ChildGaze provides a useful extension of gaze-centered datasets by supporting the study of collaborative behavior at the person level, while leaving room for further validation and expansion as the benchmark develops.

\section{Conclusion and Future Work}

This paper introduced \emph{ChildGaze}, a child-centered behavioral annotation dataset for studying collaborative behavior in naturalistic interactions. Built on the ChildPlay video collection, ChildGaze extends gaze-centered analysis by introducing two behavioral labels, \emph{collaborative} and \emph{non-collaborative}, assigned independently to each child within a frame. The dataset contains 10,641 annotated frames, 73,268 body-part annotation rows, and 13,330 child person-instances. Face, left-hand, and right-hand bounding boxes are provided for children and adults, and the annotations are organized at the row, person, and frame levels. Inter-annotator validation showed strong agreement, with a Cohen's $\kappa$ of 0.8631 for the collaboration labels and an overall mean IoU of 0.808 for the bounding-box annotations. Baseline experiments with ViT and Swin Transformer further showed that the proposed labels can be learned effectively, achieving up to 97.44\% child-person-level accuracy and 96.80\% frame-level accuracy. 

Future work will focus on expanding ChildGaze with additional videos and validating a larger portion of the dataset across more interaction scenarios. The current binary formulation can also be extended with finer behavioral categories that capture different levels or forms of collaboration. In addition, future experiments will investigate temporal video models and multimodal approaches that combine gaze, face, hand, object, and interaction cues. These extensions can provide a more complete representation of how collaborative behavior develops over time and improve the generalization of child-centered behavior understanding in naturalistic environments.

\section*{Acknowledgment}

The authors acknowledge the support and resources provided by the Bioinspired Robotics, AI, Imaging and Neurocognitive Systems (BRAINS) Laboratory, Predictive Analytics and Technology Integration (PATENT) Laboratory and IMADE (Initiative on Manufacturing Development and Education) at The University of Alabama. The authors would like to thank Harrison Wilson, Khaled Khurram, and Max Mathers for their contributions to the manual annotation of the ChildGaze dataset using Labelbox. \href{https://www.kaggle.com/datasets/penchalasindhuja/childgaze-a-benchmark-dataset}{ChildGaze} is derived from the ChildPlay-Gaze dataset created by Samy Tafasca, Anshul Gupta, and Jean-Marc Odobez, available at https://zenodo.org/records/8252535. The original material is licensed under the Creative Commons Attribution-NonCommercial 4.0 International (CC BY-NC 4.0) license. ChildGaze modifies and extends the original material through additional behavioral and collaboration annotations. The original dataset and its creators are acknowledged in accordance with the CC BY-NC 4.0 attribution requirements. The license is available at https://creativecommons.org/licenses/by-nc/4.0/.

\end{document}